\documentclass[preprint,review,12pt]{elsarticle}

\usepackage{adjustbox}
\usepackage{tabularx}
\usepackage{setspace}

\usepackage{amssymb}

\usepackage{lineno}
\usepackage{amsmath}

\usepackage{amsmath}
\usepackage{algorithm}
\usepackage[noend]{algpseudocode}

\usepackage{subcaption}

\usepackage{hhline}

\usepackage{hyperref}

\usepackage{multirow}  
\usepackage{multicol} 
\usepackage{longtable}  
\usepackage{pdflscape}  

\usepackage{enumitem}  

\usepackage{booktabs}

\usepackage{soul}
\usepackage{xcolor}

\journal{Smart Agricultural Technology}

\begin{document}

\begin{frontmatter}

\title{Using Deep Learning for Morphological Classification in Pigs with a Focus on Sanitary Monitoring}

\address[label1]{Universidade Catolica Dom Bosco, Campo Grande, Brasil}
\address[label2]{Federal University of Mato Grosso do Sul, Campo Grande, Brazil}
\address[label3]{Animal Science}

\author[label1]{Eduardo Bedin\corref{cor1}}
\ead{ra865659@ucdb.br}
\cortext[cor1]{Corresponding author}

\author[label2]{Junior Silva Souza}
\ead{junior.souza@ufms.br}

\author[label1]{Gabriel Toshio Hirokawa Higa}
\ead{gabrieltoshio03@gmail.com}

\author[label3]{Alexandre Pereira}
\ead{alepereiraszoo@gmail.com}

\author[label2]{Charles Kiefer}
\ead{charles.kiefer@ufms.br} 

\author[label1]{Newton Loebens}
\ead{newtonloebens@gmail.com}

\author[label1,label2]{Hemerson Pistori}
\ead{pistori@ucdb.br}


\begin{abstract}





The aim of this paper is to evaluate the use of D-CNN (Deep Convolutional Neural Networks) algorithms to classify pig body conditions in normal or not normal conditions, with a focus on characteristics that are observed in sanitary monitoring, and were used six different algorithms to do this task. The study focused on five pig characteristics, being these caudophagy, ear hematoma, scratches on the body, redness, and natural stains (brown or black).
The results of the study showed that D-CNN was effective in classifying deviations in pig body morphologies related to skin characteristics. 
The evaluation was conducted by analyzing the performance metrics Precision, Recall, and F-score, as well as the statistical analyses ANOVA and the Scott-Knott test.
The contribution of this article is characterized by the proposal of using D-CNN networks for morphological classification in pigs, with a focus on characteristics identified in sanitary monitoring. Among the best results, the average Precision metric of 80.6\% to classify caudophagy was achieved for the InceptionResNetV2 network, indicating the potential use of this technology for the proposed task. Additionally, a new image database was created, containing various pig's distinct body characteristics, which can serve as data for future research.



\end{abstract}



\end{frontmatter}


\section{Introduction}
\label{Intro}


Sanitary monitoring is crucial to ensure the overall health and well-being of animals in pig farms, as well as the quality and safety of the final products \cite{Brum2013Doencas}. It usually involves a number of inspections, in which one may look for common symptoms such as sneezing, coughing, diarrhea, omphalitis, and lameness \cite{ABCS2014Produco}.
The health of the herd also plays a crucial role as a factor that substantially demands time and effort. This is due to the fact that it involves applying vaccines, administering treatments and also handling mortalities. Labor issues can have a negative impact on pig production performance, particularly when employees are responsible for repairs, feed production, and manure applications, which can consume valuable time and reduce productivity. To increase production performance and decrease costs, it is essential to search for ways to increase labor efficiency through strategies such as economies of scale, automation, robotization, and rationalization \cite{MARTEL_2008} \cite{Ruguet_2011}.


From a visual standpoint, according to \cite{Barron1952Siwne}, some of the main characteristics to look for when evaluating pig health status include the color of the pig's epidermis, scratch marks or cuts, and the presence of abscesses, spots, or crusts on the skin. A healthy pig should have a rosy or blushing dermis color, an excessively white or pale dermis may indicate anemia, and reddish tones tending to purple, in turn, may indicate poor blood circulation or breathing problems. Scratch marks, cuts, signs of recent tail-bitting or aggressive behavior, as well as abscesses, skin spots, and crusts on the skin can also indicate health or behavioral issues in pigs. Therefore, regular monitoring of the visual aspect of pigs is important, since identifying and addressing potential health issues early on can help prevent disease outbreaks \cite{Carrion2020Computer} and improve the overall health of the pig herd.

The absence of injuries, wounds, lesions, and skin conditions is an indicator of good health and the absence of diseases in pigs. For instance, some diseases can cause inflammation, discoloration, and spots, resulting in abnormal pig skin color, as described in \cite{Pig_Welfare}. In this context, tail biting serves as a bodily indicator of health in pigs, as it indicates an injury caused by aggressive behavior, resulting in lesions \cite{Old_breeds}. Ear hematomas are a problem that occurs in pig herds and are caused by the rupture of blood vessels, leading to a collection of blood predominantly in the ears, management practices should be implemented to prevent their occurrence \cite{Management_Ear}. Skin lesions are frequently used as a proxy indicator of post-mixing aggressive behavior, and scratches are linked to injuries caused by aggressive behavior \cite{Accumulation}.


Inspections in the herd can be used as a tool to monitor pig health and welfare. A monitoring system can provide veterinarians and producers with
several reports about the health status of the herd, giving a quick identification of potential health issues before they worsen, resulting in better management practices, reduced treatment costs, and optimized herd performance \cite{Assessment}.
Since many diseases of interest present visual symptoms, computer vision (CV) and machine learning are powerful tools in epidemiology and disease control, especially in the context of animal production, where they can be used to assist in, or even to automate, livestock monitoring, allowing farmers and other stakeholders to quickly identify and address any issues that may arise. Another advantage is that it is a less invasive monitoring method since it avoids techniques such as radio frequency identification from chips, identification by painting pigs, microphones, and accelerometers \cite{Madonna2019Precision}.
Overall, the use of computer vision and machine learning in epidemiology and disease control has enormous potential, and there are many current research opportunities in this area \cite{Gomez2021ASystematic,Larsen2021Information}.

As per sources like \cite{Old_breeds}, modern production systems that prioritize rapid growth of pigs for meat production have significant negative impacts on animals' health, given the stressful conditions that they face in the facilities. This is also a cause of symptoms such as tail-biting, cuts, and scratches. Furthermore, studies such as the one described in \cite{Tail_New_perspect_2} state that identifying and classifying tail-biting is challenging because of the wide range of visual characteristics that damaged tails can exhibit, which can vary from minor abrasions to severe putrefaction. The behavior that leads to these injuries is also essential, as it can be sudden, strong, and obsessive, which affects the severity of the damage caused. As reported in \cite{Pig_Skin_Lesion}, the types of bruises, scratches, cuts, and tail-biting observed in pigs are directly linked to their specific behaviors.

Usually, in computer vision research applied in a pig environment, the images are captured using artificial or natural illumination and they are normally captured from an upper view, typically using RGB color space \cite{riekert2020automatically}. The camera setup, including its positioning, and the lighting conditions are critical aspects of any image-based research project, and researchers should carefully consider these factors when designing their experiments, as discussed in \cite{yang2020review}.
The use of Precision Livestock Farming (PLF) can reduce time-consuming tasks and extra labor for farmers, these technologies can help monitor health indicators, allowing for the measurement of the health status of the pig herd through parameters such as vital signs, body temperature, sounds and calls, behavioral signs, facial expressions, skin color, and weight \cite{SADEGHI_Pig_lets_Health_Sensor}.

The use of smartphones in farms is becoming an increasingly common practice, farmers adopt this kind of technology to assist in their daily activities, improve efficiency, and reduce costs \cite{SmartPhone_Farmers}. 
Some traditional methods of applying computer vision technologies require the use of computers and complex software systems, and they involve high-cost investments and personnel trained in these technologies, in this context, the use of mobile devices to capture images appears as a good solution \cite{THAPAR_PIG_SmartPhone}.
The use of mobile devices to acquire images for a Computer Vision system can reduce costs and labor associated with image acquisition. They don't require a fixed point to install the device, which allows for more flexibility in capturing images from different positions and angles and are easy to use with flexible technologies \cite{PIG_RGB_D_MObile}.
When a deep learning system achieves good results using high-resolution images captured by a mobile and low-cost device, it is an indicator that the proposed system can be developed in practical terms with few financial requirements \cite{GoPro_DL}.


The application of computational techniques for animal farming already has important precedents. Among the tasks performed, it is possible to include the identification of pigs, posture recognition and tracking \cite{XU2022AutomaticScoring} \cite{ZHU2020AutomaticRecognition} and behavioral monitoring, as well as health and well-being assessment, as discussed in \cite{brunger2020panoptic} and in \cite{Domun2019Learning}.
In the field of Computer Vision, Deep Convolutional Neural Networks (CNNs) are a type of deep learning algorithm that has become widely used. CNNs have shown good performance not only in computer vision tasks such as classification, object detection, and semantic segmentation but also in natural language processing, as discussed in \cite{ALBAWI207Understanding}. Regarding its application to pig farming, \cite{MARSOT2020PIGFACE} used CNNs for pig facial recognition with high-resolution images, and \cite{HANSEN2018Towards} used CNNs to identify 10 pig faces using standard-resolution images. 

CNNs have been used for the detection and monitoring of swine behavior patterns in various works, such as that of \cite{liu2020computer}, which studied the detection of caudophagy, and that of \cite{Domun2019Learning}, which monitored the occurrence of caudophagy with diarrhea. Other studies have focused on the recognition of feeding behavior \cite{chen2020classification}, on the monitoring of behavioral patterns in environments with varying greenhouse gas concentrations using Faster R-CNN and YoloV4 \cite{Anil2021Deep}, and on the use of Mask CNN to study pig mounting behavior \cite{Li2019MountingBehavior}. R-CNN was used by \cite{Tu2021automatic} for segmentation and automatic detection in pigs. \cite{XU2022ResearchLying} applied CNN-SVM to automate the determination of the body score of pigs with segmentation. \cite{Witte2021Evaluation} compared a Mask CNN with other neural networks for swine segmentation in the production environment.

Regarding measuring pig health through body conditions, some works focus on using computer vision techniques to measure shape and temperature as health indicators \cite{A_Novel_PIG_Body_Measure}, using segmentation to construct a multi-source pig-body multi-feature representation to analyze pig health. Other studies focus on detecting pig body posture as a health indicator \cite{riekert2020automatically} using 2D cameras. Other lines of research focus on the use of thermal images to identify high temperatures in some parts of the pig's body \cite{Dayanne_Lemos} that can be related to an injury or other health deviation.

During the authors' search, to write this paper, no related works were found about using computer vision techniques to identify and classify pig body characteristics related to health status, such as scratches, ear hematomas, caudophagy, redness, or other morphological characteristics. Based on this, the purpose of this work is to evaluate the use of D-CNN (Deep Convolutional Neural Networks) algorithms in classifying pig body morphologies as normal conditions or non-normal conditions, with a major focus on characteristics observed in sanitary monitoring. The classified characteristics are caudophagy, ear hematoma, scratches, signs of redness in the body, and natural spots, each one was tested separately. Six D-CNN architectures were employed for evaluation: DenseNet201, EfficientNet, InceptionResNetV2, InceptionV3, MobileNetV2, and Xception, along with the SGD optimizer.

Five experiments were conducted, each one corresponding to a specific morphological characteristic to be classified. For each experiment, a specific dataset was constructed, consisting of two distinct classes: one representing the normal condition and the other representing the non-normal condition.

The contribution of this article revolves around the proposed utilization of D-CNN (Deep Convolutional Neural Networks) algorithms for the morphological classification of pigs, the focus is primarily on classifying pig body characteristics observed during sanitary monitoring. Among the notable results, the InceptionResNetV2 network demonstrated an average Precision of 80.6\% to classify caudophagy, indicating the potential applicability of this technology for the proposed task. Furthermore, the article also presents the creation of a new image database containing pigs with diverse body characteristics, which can serve as valuable data for future research purposes.
In future applications, this classification approach could be used to design a system capable of identifying abnormalities in pigs, thereby assisting farmers and veterinarians in more accurately identifying these characteristics.

\section{Materials and Methods}
\label{methods}

\subsection{Image acquisition}\label{lab:dataset}

The images were collected in a pig farm located in São Gabriel do Oeste, Mato Grosso do Sul, Brazil, with animals aged around 70-180 days, weighing from 30kg to 130kg, in a period from March until July of 2022. An authorization request for the collection of the images was submitted to the Ethics Committee on Animal Use of Dom Bosco Catholic University. The authorization was granted under the certificate Nº 04/2022.

Five different datasets were constructed for this study, with each dataset dedicated to analyzing a specific morphology characteristic such as (1)caudophagy, (2)ear hematoma, (3)redness, (4)scratches, and (5)natural spots. Three characteristics refer to marks on the skin: redness, scratches, and natural spots, and two refer to the shape of the body: auricular hematoma and caudophagia. Four of these morphological characteristics refer to health problems, as mentioned by \cite{Barron1952Siwne}, \cite{Pig_Welfare}, \cite{Old_breeds}, \cite{Management_Ear}, and  \cite{Accumulation}. The analysis also included the selection of natural spots as a characteristic to check if the networks can be useful in distinguishing these natural patterns in pig bodies.

Each dataset was divided into two classes: one for normal conditions of the morphology and the other for non-normal conditions. To build these datasets, a smartphone device was used to capture pig images in a production environment. Table \ref{Table_1} displays the parameters used for image acquisition. After capturing the images, they were disposed of and divided into specific datasets based on the morphology to be analyzed. The images were captured angular to the floor where the animals were located and the place was influenced by both natural and artificial light, the presence of well-illuminated images, and poorly illuminated images with the presence of shadow or low light incidence. The captures took place in the morning, between 8:00 a.m. and 10:00 a.m., as well as in the afternoon, between 1:00 p.m. and 3:30 p.m. The animals were housed in buildings with natural ventilation or climate-controlled buildings.


\begin{table}[h!] 
    \centering
    \caption{Parameters of Device and Environment in Image Capture Procedure}
    \begin{tabular}{|c|c|}
     \hline    
     Parameter  &  Value \\ [0.5ex]
     \hline
     \hline
     Device   & Moto e(6i) \\
     \hline
     Focal Distance & 3.7mm \\
     \hline 
     Original Resolution & 2340 x 4160 pixels \\
     \hline
     Acquisition & Manual \\
     \hline
     Height & 1.00m - 1.5m \\
     \hline
     Acquisition Angle & Angular \\
     \hline
     Illumination & Natural and Artificial \\
     \hline
    \end{tabular}
    
    \label{Table_1}
\end{table}

\begin{figure}
     \centering
     \begin{subfigure}[]{0.3\textwidth}
         \centering
         \includegraphics[width=\textwidth,height=3cm]{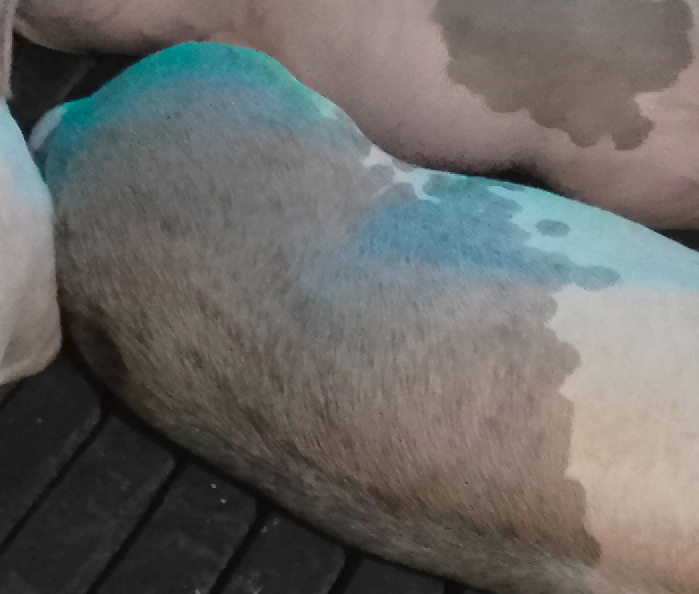}
         \caption{}
         \label{fig:mancha}
     \end{subfigure}
     \begin{subfigure}[]{0.3\textwidth}
         \centering
         \includegraphics[width=\textwidth,height=3cm]{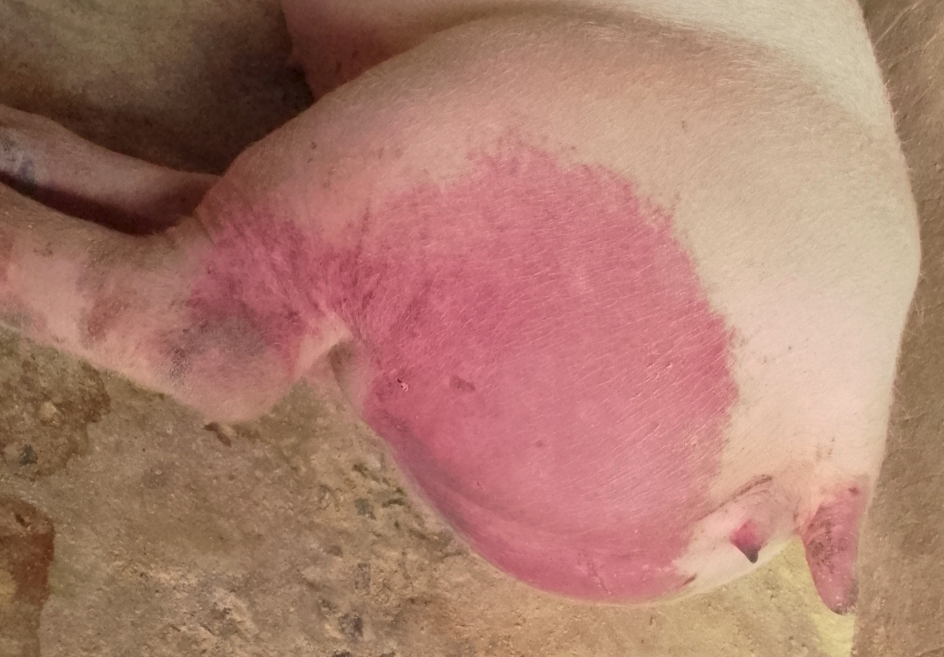}
         \caption{}
     \end{subfigure}
     \begin{subfigure}[]{0.3\textwidth}
         \centering
         \includegraphics[width=\textwidth,height=3cm]{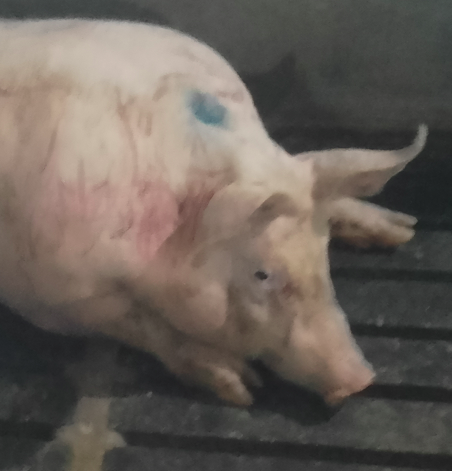}
         \caption{}
     \end{subfigure}
     \begin{subfigure}[]{0.3\textwidth}
         \centering
         \includegraphics[width=\textwidth,height=3cm]{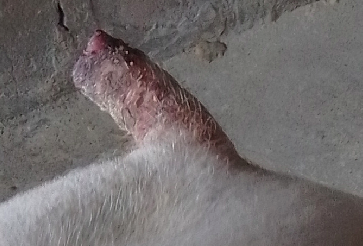}
         \caption{}
     \end{subfigure}
     \begin{subfigure}[]{0.3\textwidth}
         \centering
         \includegraphics[width=\textwidth,height=3cm]{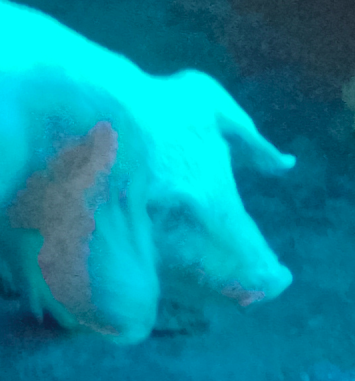}
         \caption{}
     \end{subfigure}
        \caption{Examples of pig images that are used in the five datasets. Spotted Pig (a), Pig with Redness (b), Scratched Pig(3), Pig Tail with Caudophagy (4) and One Example of Pig with Ear Hematoma (5)}
        \label{detalhe_pig}
\end{figure}

As the aim of this paper is to classify the proposed morphologies as normal or non-normal conditions, after image acquisition, and the images disposed of and divided into specific datasets based on the morphology to be analyzed, the dataset-building process was finalized by cropping the original images to highlight the relevant morphology. This was done to eliminate unnecessary information contained in the images and focus solely on the respective morphology that the network should classify in each task.
Figure \ref{detalhe_pig} shows samples of the images that constitute each dataset. The five datasets are comprised of 715 images. Table \ref{table:num_pictures} shows the number of images that compose each dataset used in each experiment. It is important to emphasize that each dataset was used separately in each experiment, to classify the specific morphology.


\begin{table}[h!]
    \centering
    \caption{Quantity of images in each morphological dataset. This table columns represent the groups, Number of Images, Number of Anomalous, and Number of Normal Images in each dataset.}
    \begin{tabular}{|c|c|c|c|c|}
     \hline    
     Group  &  \#Img  & Anomalous & Percentage & N \\
     \hline
     \hline
     Caudophagy & 182 & 90 & 49.46\% & 92  \\
     \hline
     Ear Hematoma & 143 & 64 & 44.76\% & 79 \\
     \hline 
     Spots & 212 & 139 & 65.57\% & 73  \\
     \hline
     Scratches & 124 & 51 & 41.13\% & 73  \\
     \hline
     Redness & 60 & 30 & 50\% & 30  \\
     \hline
    \end{tabular}
    
    \label{table:num_pictures}
\end{table}


\subsection{Neural Networks}\label{lab:networks}

To develop this work, six different kinds of D-CNN and one optimizer were used for each morphology to be classified. The purpose was to analyze the capacity of each proposed network to classify each morphology in normal or abnormal conditions.
The networks used were EfficientNet, Xception, DenseNet201, InceptionResNetV2, InceptionV2, and MobileNetV2. Stochastic Gradient Descent (SGD) was used for optimization. Table \ref{tab:num_params} shows the standard number of parameters of each neural network\footnote{As stated here: \url{https://keras.io/api/applications/}.}.

Through the parameters, also called weights, which the algorithm learns during the training phase, the difference between Neural Networks can be analyzed by the number of parameters. When a Neural Network has more parameters, it has the capacity to learn more complex patterns. However, networks with fewer parameters use less computational resources. In this context, the appropriate choice of Neural Networks can be determined by finding a good balance between the number of parameters and computational cost \cite{LeCun} \cite{Goodfellow-et-al-2016}.


{\renewcommand{\arraystretch}{1.2}%
\begin{table}[h!]
    \centering
    \caption{Number of parameters of each D-CNN used in the five experiments }
    \begin{tabular}{|c|c|}
        \hline
        Architecture & \# of parameters \\
        \hline \hline
        EfficientNet & 19.5M \\
        \hline
        Xception & 22.9M \\
        \hline
        DenseNet201 & 20.2M \\
        \hline
        InceptionResNetV2 & 55.9M \\
        \hline
        InceptionV3 & 23.9M \\
        \hline
        MobileNetV2 & 3.5M \\
        \hline
    \end{tabular}
    
    \label{tab:num_params}
\end{table}}

The EfficientNet architecture is based on a scaling method that simultaneously increases the depth, width, and resolution of the network, which allows it to achieve higher accuracy with fewer parameters and computational resources compared to other state-of-the-art models \cite{Tan2020Efficient}. Xception is a deep neural network architecture that was proposed as an alternative to the traditional Inception architecture for image classification tasks. Its architecture replaces the standard Inception modules, which perform convolutions with a mixture of filter sizes, with depthwise separable convolutions. The use of depthwise separable convolutions in Xception results in a significant reduction in the number of parameters compared to the traditional Inception architecture~ \cite{Francois_Xception_17}.

DenseNet201 is a deep neural network architecture that belongs to the family of DenseNets, which were designed to address the issue of vanishing gradients in deep neural networks. Specifically, DenseNet201 is a very deep neural network architecture that has 201 layers, hence the name. The architecture consists of several dense blocks, where each dense block contains multiple convolutional layers that are connected to all subsequent layers in the block \cite{Huang2018Densely}. InceptionResNetV2 is a deep neural network architecture that combines the Inception and ResNet architectures, this network combines these two architectures by using Inception modules with residual connections each Inception module has a residual connection that skips over the module and directly connects the input to the output of the module \cite{InceptionResnetV2_2016}.

InceptionV3 is a deep neural network architecture that is designed for image classification tasks, the core idea behind the Inception architecture is to use convolutional layers with filters of multiple sizes at each layer. This allows the network to capture features at different scales and resolutions while minimizing the computational cost of the network \cite{InceptionV2_2018}. On the other hand, MobileNetV3 is another convolutional neural network architecture specifically designed for mobile devices, such as smartphones. It was developed using a combination of network hardware architecture (NAS) and the NetAdapt algorithm. The NAS approach involves the use of reinforcement learning to search for optimal neural network architectures, while NetAdapt is an algorithm used to adapt neural networks to different hardware constraints \cite{MobileNetV3}.

The optimizer used was SGD (Stochastic Gradient Descent), which is a popular optimization algorithm used in deep learning to train neural networks \cite{shim2023modified}. The basic idea behind SGD is to take small steps in the direction of the steepest descent of the loss function, i.e., the direction in which the loss function decreases the fastest. This is achieved by computing the gradient of the loss function with respect to the model parameters for a randomly selected subset of the training data, known as a mini-batch. The model parameters are then updated based on this gradient \cite{SGD_2013}.

\subsection{Experimental setup}\label{lab:experimental}

The experimental setup consists of performing five experiments using the same six D-CNNs and the SGD optimizer mentioned in Subsection \ref{lab:networks}. Each experiment corresponds to a specific morphology from the datasets mentioned in Section \ref{lab:dataset}: Ear Hematoma, Caudophagy, Scratches, Redness, and Natural Spots. This procedure aims to evaluate the applicability of these networks in classifying images of pig morphologies as normal or non-normal conditions.

The Deep Neural Networks\footnote{In this work, the TensorFlow \cite{tensorflow2015-whitepaper} framework and the Keras API \cite{chollet2015keras} were used to train the neural networks.} listed above were tested in a 10-fold stratified cross-validation strategy, and used the Data Augmentation technique to improve the images samples during the training phase. In each experiment, 30\% of the training images were used for validation. The values of hyperparameters were adjusted according to experimental tests during the development phase of this work, and the range of values that showed good results were presented in Table \ref{Table_4}. These values were the same for all neural networks.


\begin{table}[h!]
\begin{center}
    
    \caption{Hyper-parameters used in the five experiments and their values}
    \label{Table_4}
    \begin{tabular}{|p{4,0cm}|p{4,0cm}|}
    \hline
    Hyperparameters                    & Value                   \\ [0.5ex]
    \hline
    Learning Rate                      & 0.001                   \\ 
    \hline
    Learning Rate Decay                & 0.0001                  \\ 
    \hline
    Bacth Size                         & 8                       \\ 
    \hline
    Epochs 200                         & 200                     \\ 
    \hline
    Patience Percentage                & 15                      \\ 
    \hline
    Validation Set                     & 30                      \\ 
    \hline
    Dropout                            & 0.5                     \\ 
    \hline
    \multirow{6}{*}{Data Augmentation} & Flip = True             \\ \cline{2-2} 
    & Fill Mode= "Nearest"    \\ \cline{2-2} 
                                   & Zoom = 10               \\ \cline{2-2} 
                                   & Height Shift Range= 0.5 \\ \cline{2-2} 
                                   & Widht Shift Range= 0.5  \\ \cline{2-2} 
                                   & Rotation = 90           \\ \hline
    \end{tabular}
    \end{center}
\end{table}


Three common classification metrics were used to evaluate the algorithms: Precision, Recall, and F-score. Precision is used as a measure of the proportion of correctly predicted positive instances out of all instances, focusing on the accuracy of positive predictions. Recall measures the proportion of correctly predicted positive instances out of the actual positive instances, focusing on the model's ability to find all positive instances. The F-score metric is the harmonic mean of Precision and Recall. This metric is useful when the costs of false positives and false negatives differ. Box plots were generated from the metrics Precision, Recall, and F-Score. Grad-CAM scores were also generated and used in the analysis of the results. This technique enables the visualization of visual maps that highlight where the algorithm focuses its attention in the classification task. After running the model and saving the results, an ANOVA hypothesis test was applied with a threshold of 5\% in the three metrics analyzed. The Scott-Knott clustering test was used post-hoc when the ANOVA results were significant\footnote{Both the ANOVA and the Scott-Knott clustering test were performed in the R Programming Language \cite{team2022r,jelihovschi2014scottknott}.}.


\section{Results and discussion}

The replicability of the results of this paper is delimited by the procedure and conditions of image acquisitions described in subsection 2.1, and to the networks, and hyperparameters presented in sections 2.2 and 2.3. It is important to emphasize that the primary goal of the aforementioned procedure is to analyze the capacity of the D-CNN (Deep Convolutional Neural Network) in classifying some specific pig morphologies as normal or abnormal.

\begin{figure}
    \centering
    \includegraphics[scale=0.70]{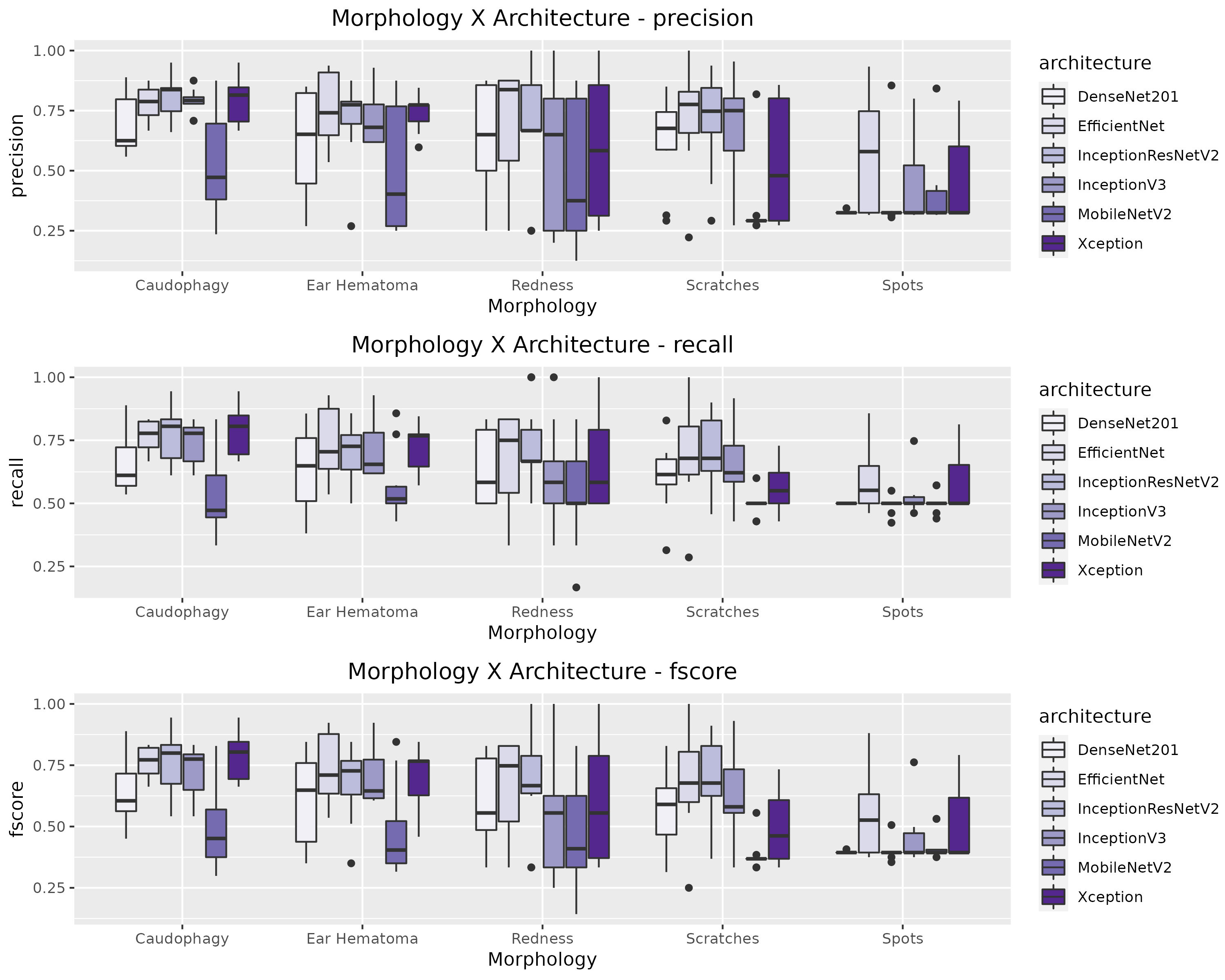}
    \caption{Boxplot of metrics performance networks in the function of the morphology that was classified in each experiment}
    \label{fig:Box_plot_todos}
\end{figure}

\begin{figure}
     \centering
     \begin{subfigure}[]{0.35\textwidth}
         \centering
         \includegraphics[width=\textwidth,height=4cm]{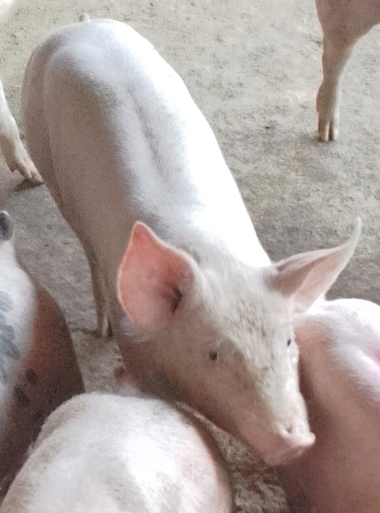}
         \caption{}
         \label{fig:porco_limpo}
     \end{subfigure}
     \begin{subfigure}[]{0.35\textwidth}
         \centering
         \includegraphics[width=\textwidth,height=4cm]{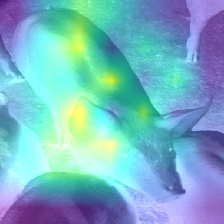}
         \caption{}
     \end{subfigure}
        \caption{GradCAM analisys of No-Spot that was wrong Classificated.}
        \label{fig:gradcam_porco_limpo}
\end{figure}

\begin{figure}
     \centering
     \begin{subfigure}[]{0.35\textwidth}
         \centering
         \includegraphics[width=\textwidth,height=4cm]{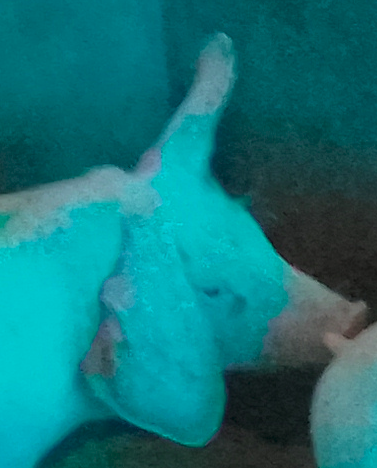}
         \caption{}
         \label{fig:_Slowen_Ear}
     \end{subfigure}
     \begin{subfigure}[]{0.35\textwidth}
         \centering
         \includegraphics[width=\textwidth,height=4cm]{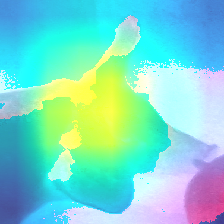}
         \caption{}
     \end{subfigure}
        \caption{GradCAM of ear hematoma that was correctly classified.}
        \label{fig:Grad_Ear}
\end{figure}

\begin{figure}
     \centering
     \begin{subfigure}[]{0.35\textwidth}
         \centering
         \includegraphics[width=\textwidth,height=4cm]{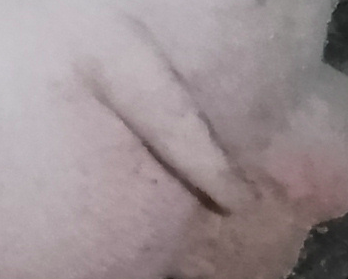}
         \caption{}
         \label{fig:_Not_cannib}
     \end{subfigure}
     \begin{subfigure}[]{0.35\textwidth}
         \centering
         \includegraphics[width=\textwidth,height=4cm]{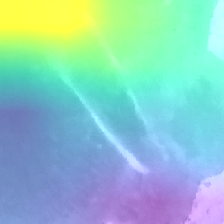}
         \caption{}
     \end{subfigure}
        \caption{GradCAM of normal tail mistakenly classified as tail with caudophagy.}
        \label{fig:Grad_Cau}
\end{figure}

\begin{figure}
     \centering
     \begin{subfigure}[]{0.35\textwidth}
         \centering
         \includegraphics[width=\textwidth,height=3cm]{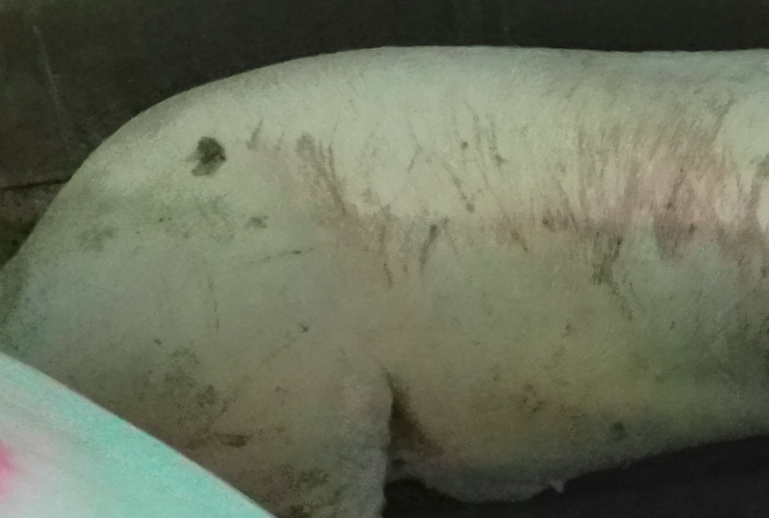}
         \caption{}
         \label{fig:Scrat_scrat}
     \end{subfigure}
     \begin{subfigure}[]{0.35\textwidth}
         \centering
         \includegraphics[width=\textwidth,height=3cm]{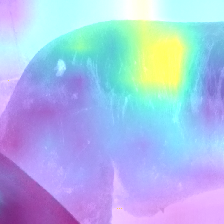}
         \caption{}
     \end{subfigure}
        \caption{GradCAM of scratches in pig body that was correctly classified.}
        \label{fig:Grad_Scrat}
\end{figure}

\begin{figure}
     \centering
     \begin{subfigure}[]{0.35\textwidth}
         \centering
         \includegraphics[width=\textwidth,height=6cm]{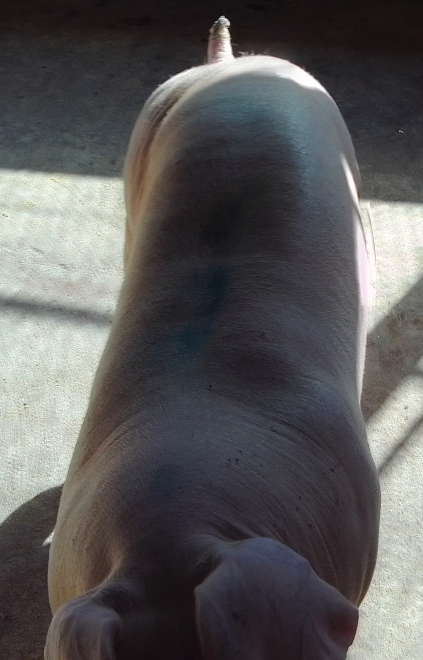}
         \caption{}
         \label{fig:No_red_red}
     \end{subfigure}
     \begin{subfigure}[]{0.35\textwidth}
         \centering
         \includegraphics[width=\textwidth,height=6cm]{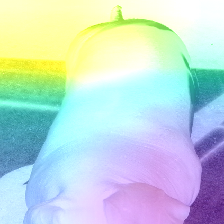}
         \caption{}
     \end{subfigure}
        \caption{GradCAM of normal pig skin color that was mistakenly classified as redness.}
        \label{fig:Grad_Red}
\end{figure}

The boxplots for each performance metric are shown in Figure \ref{fig:Box_plot_todos}. For each metric, boxplots representing each architecture are grouped by morphology. Tables \ref{table:precision}, \ref{table:recall} and \ref{table:fscore} show statistics for Precision, Recall, and F-score, respectively, for each architecture, also grouped by morphology. Next to the mean values, the results of the Scott-Knott clustering test are indicated. The ANOVA results can be found in Table \ref{table:Anova_Test}. Figures \ref{fig:gradcam_porco_limpo}, \ref{fig:Grad_Ear}, \ref{fig:Grad_Cau}, \ref{fig:Grad_Scrat} and \ref{fig:Grad_Red} show the GradCam comparison of Spots, Ear Hematoma, Caudophagy, Scratches, and Redness.

The ANOVA test indicates that for scratching and caudophagy the null hypothesis is to be rejected for each one of the metrics - Precision, Recall, and F-score. For auricular hematoma, the ANOVA results for Precision and F-score were also significant, but the ANOVA result for Recall was marginally insignificant ($p=0.0554$). In this case, the Scott-Knott test was still applied and yielded different clusters. For spots, the null hypothesis is to be accepted for Precision, with a fully insignificant ANOVA result ($p=0.1060$), but not for Recall and F-score. As for redness, the ANOVA results were not significant for any of the metrics.

{\renewcommand{\arraystretch}{1.2}%
\begin{table}[h!]

\caption{ANOVA Summary for Each Experimental Test}
\label{table:Anova_Test}
\centering
\resizebox{\linewidth}{!}{
\begin{tabular}[t]{|c|c|c|c|}
\toprule
\hline

Morphology & Precision & Recall & Fscore \\
\hline \hline
Scratches  & 0.0010 & 0.0090 & 0.0003  \\
\hline
Ear hematoma & 0.0218 & 0.0554  &  0.0075  \\
\hline
Caudophagy & 1.04e-05 & 7.19e-06  &  4.67e-06  \\
\hline
Spots & 0.1060 & 0.0431  &  0.0243  \\
\hline
Redness & 0.5910  &  0.3140 & 0.2840   \\

\hline
\bottomrule
\end{tabular}}
\end{table}
}



Figure \ref{fig:Box_plot_todos} shows that many of the larger IQR values were observed for the redness morphology, which suggests greater heterogeneity in this dataset since it was divided in folds, but its reduced size can be considered responsible for at least part of the great variability in these results, since it means that there is less data available for training and, most importantly for this specific point, for testing, in each one of the ten folds. On the other hand, the spots morphology, which has the largest dataset, presented the lowest observed averages, with small interquartile ranges. 

For the natural spots, the ANOVA results for Precision were not significant, and, as Tables \ref{table:precision}, \ref{table:recall} and \ref{table:fscore} show, EfficientNet and Xception networks achieved the higher performance measured by Recall and F-score. Possible reasons for spots being of more difficult identification for the neural networks are that these natural spots are more likely to be confused with dirt and other environment-related conditions (\textit{e.g.} shadows) in virtue of their color (dark-grayish, as opposed to the more vibrant reddish tones).
Admittedly, Figure \ref{fig:porco_limpo} does not seem to follow this logic, and one could argue that the neural network is giving more importance to the actually spotted pig on the left. On the other hand, Figure \ref{fig:gradcam_porco_limpo}, which shows the corresponding GradCAM, suggests that the non-spotted pig does influence the prediction, and it is possible to support the presented argument by blaming the light conditions that are causing shadows that resemble spots.

More generally, during the evaluation of network errors in image classification, certain characteristics were identified. For instance, when using a morphology-based format to identify features like caudophagy and ear hematoma, the D-CNN model tended to misclassify images with poorly-defined formats or with environment colors similar to the body's tone. In addition, the model struggled with distinguishing between normal characteristics and mild deformations in such morphologies, leading to incorrect classifications. Figure \ref{fig:Grad_Ear} is an example of right Ear Hematoma classification, when is possible to evaluate that the network focused on the deformed ear to classify the image correctly, and \ref{fig:Grad_Cau} is an example of wrong Caudofagy classification, when is possible to observe that the network focused in an upper side of the tail and classified it as a caudophagy but it isn't.





\onehalfspacing

{\renewcommand{\arraystretch}{0.6}
\begin{table}

\caption{Precision statistics: The table shows the networks that presented the best results.}
\label{table:precision}
\centering

\resizebox{\textwidth}{!}{
\begin{tabular}[t]{|c|c|c|}
\toprule
\multicolumn{3}{c}{\textbf{Scratches}} \\
\hline
Architecture & Median (IQR) & Mean SK (SD)\\
\hline
EfficientNet & \textbf{0.7757 (0.1715)} & \textbf{0.7178 a ($\pm 0.2116$)}\\
\hline

\multicolumn{3}{c}{\textbf{Ear hematoma}} \\
\hline
\hline
EfficientNet & 0.7411 (0.2614) & \textbf{0.7614 a ($\pm 0.1477$)}\\
InceptionResNetV2 & \textbf{0.7744 (0.0926)} & 0.7117 a ($\pm 0.1719$)\\
\hline

\multicolumn{3}{c}{\textbf{Caudophagy}} \\
\hline
\hline
InceptionResNetV2 & \textbf{0.8375 (0.0965)} & \textbf{0.8063 a ($\pm 0.0954$)}\\
\hline

\multicolumn{3}{c}{\textbf{Spots}} \\
\hline
\hline
EfficientNet & \textbf{0.5795 (0.4227)} & \textbf{0.5646 ($\pm 0.2374$)}\\
\hline

\multicolumn{3}{c}{\textbf{Redness}} \\
\hline
\hline
EfficientNet & \textbf{0.8375 (0.3333)} & \textbf{0.6925 ($\pm 0.1835$)}\\
\hline
\bottomrule
\end{tabular}}

\end{table}
}

\doublespacing

{\renewcommand{\arraystretch}{0.8}
\begin{table}

\caption{Recall statistics. The table shows the networks that presented the best results.}
\label{table:recall}
\centering
\resizebox{\linewidth}{!}{
\begin{tabular}[t]{|c|c|c|}
\toprule
\multicolumn{3}{c}{\textbf{Scratches}} \\
\hline
Architecture & Median (IQR) & Mean SK (SD)\\
\hline
EfficientNet & \textbf{0.6786 (0.1905)} & 0.6876 a ($\pm 0.1889$)\\
\hline
InceptionResNetV2 & \textbf{0.6786 (0.2000)} & \textbf{0.7029 a ($\pm 0.1578$)}\\
\hline

\multicolumn{3}{c}{\textbf{Ear hematoma}} \\
\hline
\hline
EfficientNet & 0.7046 (0.2381) & \textbf{0.7362 a ($\pm 0.1417$)}\\
\hline
Xception & \textbf{0.7678 (0.1279)} & 0.7183 a ($\pm 0.0927$)\\
\hline

\multicolumn{3}{c}{\textbf{Caudophagy}} \\
\hline
\hline
Xception & \textbf{0.8056 (0.1540)} & \textbf{0.7909 a ($\pm 0.0985$)}\\
\hline
InceptionResNetV2 & \textbf{0.8056 (0.1540)} & 0.7717 a ($\pm 0.1157$)\\
\hline

\multicolumn{3}{c}{\textbf{Spots}} \\
\hline
\hline
EfficientNet & \textbf{0.5516 (0.1483)} & \textbf{0.5927 a ($\pm 0.1358$)}\\
\hline

\multicolumn{3}{c}{\textbf{Redness}} \\
\hline
\hline
EfficientNet & \textbf{0.7500 (0.2916)} & 0.6833 ($\pm 0.1834$)\\
\hline
InceptionResNetV2 & 0.6667 (0.1249) & \textbf{0.7166 ($\pm 0.1765$)}\\
\hline
\bottomrule
\end{tabular}}
\end{table}
}

{\renewcommand{\arraystretch}{0.8}
\begin{table}

\caption{F-Score statistics. The table shows the networks that presented the best results.}
\label{table:fscore}
\centering
\resizebox{\linewidth}{!}{
\begin{tabular}[t]{|c|c|c|}
\toprule
\multicolumn{3}{c}{\textbf{Scratches}} \\
\hline
Architecture & Median (IQR) & Mean SK (SD)\\
\hline
EfficientNet & \textbf{0.6773 (0.2052)} & 0.6788 a ($\pm 0.2003$) \\
\hline
InceptionResNetV2 & \textbf{0.6773 (0.2036)} & \textbf{0.6852 a ($\pm 0.1818$)} \\
\hline

\multicolumn{3}{c}{\textbf{Ear hematoma}} \\
\hline
\hline
EfficientNet & 0.7098 (0.2430) & \textbf{0.7347 a ($\pm 0.1447$)} \\
\hline
Xception & \textbf{0.7664 (0.1420)} & 0.6976 a ($\pm 0.1241$) \\
\hline

\multicolumn{3}{c}{\textbf{Caudophagy}} \\
\hline
\hline
Xception & \textbf{0.8039 (0.1516)} & \textbf{0.7890 a  ($\pm 0.09949$)} \\
\hline

\multicolumn{3}{c}{\textbf{Spots}} \\
\hline
\hline
EfficientNet & \textbf{0.5262 (0.2378)} & \textbf{0.5494 a ($\pm 0.1779$)} \\
\hline

\multicolumn{3}{c}{\textbf{Redness}} \\
\hline
\hline
EfficientNet & \textbf{0.7476 (0.3080)} & 0.6587 ($\pm 0.2074$)\\
\hline
InceptionResNetV2 & 0.6667 (0.1527) & \textbf{0.6787 ($\pm 0.2287$)}\\
\hline
\bottomrule
\end{tabular}}
\end{table}
}




When it comes to morphologies based on skin characteristics, the success of the networks in correctly classifying images depends on the contrast between the pig's body and the background. However, the networks often make mistakes in classifying these morphologies, particularly when the images have poorly-defined contours, details close to the ground or wall, and weak contrast between the foreground and background. These factors can make it challenging for the network to distinguish between normal and abnormal skin characteristics, resulting in incorrect classifications.
Interestingly, for scratches, the density of the scratches seems to play a significant role in the network's ability to identify them. This indicates that the network may perform better at identifying scratches that are more prominent or visible. Figure \ref{fig:Grad_Scrat} is an example of correct Scratches classification, it is possible to observe that due to the high density of scratches on the pig's body, the network focuses on a specific region and makes the correct classification. On the other hand, in Figure \ref{fig:Grad_Red}, we see an example of incorrect Redness classification. In this case, the image provided to the network features an entirely white pig, but the network focuses on a shiny part and misclassifies it as redness, which is not accurate.


Upon observing the results of the Scott-Knott test, MobileNetV2 was clustered in group 'b' in 4 out of the 5 morphologies studied, with the exception of the redness morphology. This network presented the smallest means and medians in each situation, including redness, which can be attributed to its design that utilizes a reduced number of parameters, as can be seen in Table \ref{tab:num_params}. This architectural option was taken in order to achieve high performance on devices with smaller computational capabilities. This is relevant for the purposes of a future system since it suggests that smaller networks perform poorly on this task, but more tests with other smaller and newer networks may show otherwise. Also, on the other hand, using bigger neural networks does not necessarily lead to better performance in the present problem, since usually there was no significant difference between EfficientNet, the second smallest neural network tested by us, and InceptionResNetV2, the biggest one. In all the cases where a significant difference was observed, EfficientNet had the upper hand.


\section{Conclusion}

In this work, we studied the possibility of applying neural networks to the classification of normal from abnormal morphological characteristics in pigs. Turning possible to classify these characteristics is important for possible applicability in sanitary monitoring, which can help to improve animal health and welfare through the possibility of a better way to classify pig body morphology conditions. The results indicate that the proposed use is feasible, but the performance is still below necessary for the considered morphological characteristics: caudophagy, ear hematomas, redness, scratches, and spots.
Given the scarcity of references on the specific subject, it becomes notable to conduct further research in this area. Such research could be centered on studying and developing algorithms that can more accurately detect and classify abnormalities in swine bodies, which could be indicative of poor health. Moreover, the research could delve into exploring additional morphologies that may signal a deviation from good health.

\section{Declaration of competing interest}

The authors declare that they have no known competing financial
interests or personal relationships that could have appeared to influence the work reported in this paper.

\section*{Acknowledgments}

Some of the authors were awarded scholarships from the Brazilian National Council of Technological and Scientific Development (CNPQ), and Foundation University of Technology and Sciences Enterprise (Fundatec), and University Higher Education Personnel Improvement Coordination (CAPES). This work has received infrastructure support from Dom Bosco Catholic University. We would also like to thank NVIDIA for providing the Titan X GPUs used in the experiments.








\end{document}